\documentclass[10pt,twocolumn,letterpaper]{article}

\usepackage{cvpr}              

\usepackage{pgfplots}
\usepackage{pifont}
\pgfplotsset{compat=1.18}
\usepackage{xcolor}
\definecolor{baselineColor}{RGB}{54,104,191} 
\definecolor{oursColor}{RGB}{203,55,62}      

\definecolor{cvprblue}{rgb}{0.21,0.49,0.74}
\usepackage[pagebackref,breaklinks,colorlinks,allcolors=cvprblue]{hyperref}
\usepackage{dirtytalk}
\usepackage{xcolor}
\usepackage{multirow}
\usepackage{booktabs}
\usepackage{pifont}
\usepackage[table]{xcolor}
\usepackage{makecell}
\usepackage{etoolbox}

\makeatother

\def\paperID{****} 
\def\confName{CVPR}
\def\confYear{2026}

\title{Changes in Real Time: Online Scene Change Detection with Multi-View Fusion}

\author{Chamuditha Jayanga Galappaththige$^{1,2}$ \hspace{10pt} Jason Lai$^{3}$ \hspace{10pt}  Lloyd Windrim$^{2,4}$ \\ Donald Dansereau$^{2,3}$  \hspace{10pt}   Niko S\"underhauf$^{1,2}$ \hspace{10pt}  Dimity Miller$^{1,2}$ \\
$^1$QUT Centre for Robotics \hspace{1pt} $^2$ARIAM \hspace{1pt} $^3$ACFR, University of Sydney \hspace{1pt} $^4${Abyss Solutions}
\\
{\tt\small \{chamuditha.galappaththige, d24.miller\}@.qut.edu.au}
}

\begin{document}
\maketitle

\begin{abstract}
    Online Scene Change Detection (SCD) is an extremely challenging problem that requires an agent to detect \emph{relevant} changes on the fly while observing the scene from unconstrained viewpoints. Existing online SCD methods are significantly less accurate than offline approaches. We present the first online SCD approach that is pose-agnostic, label-free, and ensures multi-view consistency, while operating at over 10 FPS and achieving new state-of-the-art performance, surpassing even the best offline approaches.
    Our method introduces a new self-supervised fusion loss to infer scene changes from multiple cues and observations, PnP-based fast pose estimation against the reference scene, and a fast change-guided update strategy for the 3D Gaussian Splatting scene representation. Extensive experiments on complex real-world datasets demonstrate that our approach outperforms both online and offline baselines. Code is available at \href{https://chumsy0725.github.io/O-SCD/}{https://chumsy0725.github.io/O-SCD/}.
\end{abstract}

\section{Introduction}
\label{sec:intro}


Detecting changes in a scene is an essential task in scene understanding, with numerous applications in environmental monitoring~\cite{taneja2011image}, infrastructure inspection~\cite{han2021change}, and damage assessment~\cite{sakurada_change_2015}. Scene change detection (SCD) is especially challenging in the context of robotics, where an agent observes the scene from unconstrained and independent viewpoints when re-visiting it after some time, while having to discern relevant (e.g. object movement) and irrelevant changes (e.g. caused by shadows, or reflections).

To address these challenges, recent approaches~\cite{galappaththige2025multi,lu20253dgs,jiang2025gaussian,zhou2023pad,kruse2024splatpose} leverage photorealistic 3D scene representations like Neural Radiance Fields~\cite{mildenhall2021nerf} (NeRF) and 3D Gaussian Splatting~\cite{kerbl20233d} (3DGS) to enable \emph{pose-agnostic} SCD from unconstrained viewpoints. 
Complementary efforts~\cite{galappaththige2025multi,kim2025towards,cho2025zero} explore \emph{label-free} SCD to remove the reliance on costly and labor-intensive human-labeled changes, improving robustness under domain and data distribution shifts.

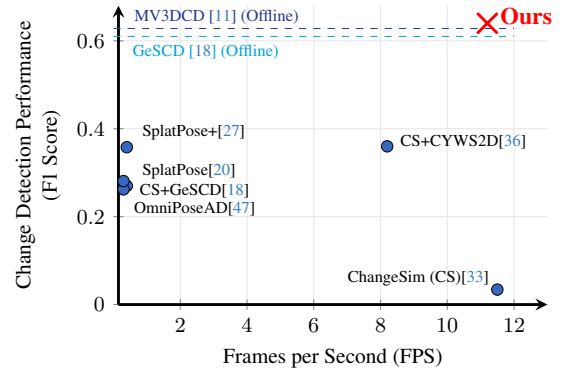
\begin{figure}[t]
  \centering
  \scalebox{0.9}{
  \begin{tikzpicture}
    \begin{axis}[
      width=0.95\linewidth,
      height=6cm,
      xlabel={Frames per Second (FPS)},
      ylabel style={align=center},
      ylabel={Change Detection Performance\\(F1 Score)},
      xmin=0.15, xmax=13,
      ymin=0.0, ymax=0.68,
      grid=both,
      grid style={line width=.2pt, draw=gray!20},
      tick label style={font=\small},
      label style={font=\small},
      axis background/.style={fill=none},
      axis line style={very thick},
      axis x line=bottom,
      axis y line=left,
      clip=false
    ]

      \addplot[
        only marks,
        mark=*,
        mark options={scale=1.2, draw=black, fill=baselineColor}
      ] coordinates {
        (11.5,0.034)  
        (8.2,0.360)   
        (0.4,0.270)   
        (0.3,0.262)   
        (0.3,0.281)   
        (0.4,0.358)   
      };

      \addplot[
        only marks,
        mark=x,
        very thick,
        color=red,
        mark options={scale=3}
      ] coordinates {(11.2,0.64)};


      \addplot[
        domain=0:12,
        samples=2,
        ultra thin,
        color=cyan!5!blue,
        dashed
      ] {0.628};
      \node[anchor=south east, font=\scriptsize, text=cyan!5!blue]
        at (axis cs:5.8,0.62) {MV3DCD~\cite{galappaththige2025multi} (Offline)};

      \addplot[
        domain=0:12,
        samples=4,
        ultra thin,
        color=cyan!90!blue,
        dashed
      ] {0.61};
      \node[anchor=south east, font=\scriptsize, text=cyan!90!blue]
        at (axis cs:5.2,0.54) {GeSCD~\cite{kim2025towards} (Offline)};

      \node[anchor=east, font=\scriptsize, yshift=5pt] at (axis cs:11.5,0.034) {ChangeSim (CS)\cite{park2021changesim}};
      \node[anchor=west, font=\scriptsize, yshift=2pt, xshift=2pt] at (axis cs:8.2,0.360) {CS+CYWS2D\cite{cyws2d}};
      \node[anchor=west, font=\scriptsize, yshift=-2pt, xshift=2pt] at (axis cs:0.4,0.270) {CS+GeSCD\cite{kim2025towards}};
      \node[anchor=west, font=\scriptsize, yshift=-8pt, xshift=1pt] at (axis cs:0.3,0.262) {OmniPoseAD\cite{zhou2023pad}};
      \node[anchor=west, font=\scriptsize, yshift=8pt, xshift=3pt] at (axis cs:0.4,0.261) {SplatPose\cite{kruse2024splatpose}};
      \node[anchor=west, font=\scriptsize, yshift=10pt, xshift=3pt] at (axis cs:0.4,0.338) {SplatPose+\cite{liu2024splatpose+}};
      \node[anchor=west, font=\scriptsize, text=red, font=\bfseries, yshift=3pt, xshift=2pt] at (axis cs:11.2,0.64) {Ours};

    \end{axis}
  \end{tikzpicture}
  }
  \vspace{-5pt}
  \caption{
  Our \emph{online} scene change detection method establishes a new state of the art, detecting changes more reliably than all prior methods, including the strongest \emph{offline} baselines. It operates at a runtime comparable to the fastest online approaches while achieving substantially higher F1 scores. 
  }
  \label{fig:f1_fps_online}
\end{figure}

Despite recent advances, state-of-the-art (SOTA) SCD methods~\cite{kim2025towards, lu20253dgs, galappaththige2025multi} are confined to an \emph{offline} setting, where both pre- and post-change observations are available prior to inference. In contrast, online change inference---referred to as \emph{online SCD}---detects changes on the fly as new images are acquired during a scene revisit, without access to future observations. This setting is critical for real-time decision-making and intervention in embodied and robotic systems. As in prior online approaches~\cite{liu2024splatpose+, kruse2024splatpose, park2021changesim, park2021changesim}, we assume that the scene remains static within each revisit. As shown in Fig.~\ref{fig:f1_fps_online}, existing online SCD methods exhibit a substantial accuracy gap compared to SOTA offline approaches. Moreover, many fail to sustain real-time performance, limiting their practical applicability.

\begin{figure*}[t]
  \centering
   \includegraphics[width=\linewidth]{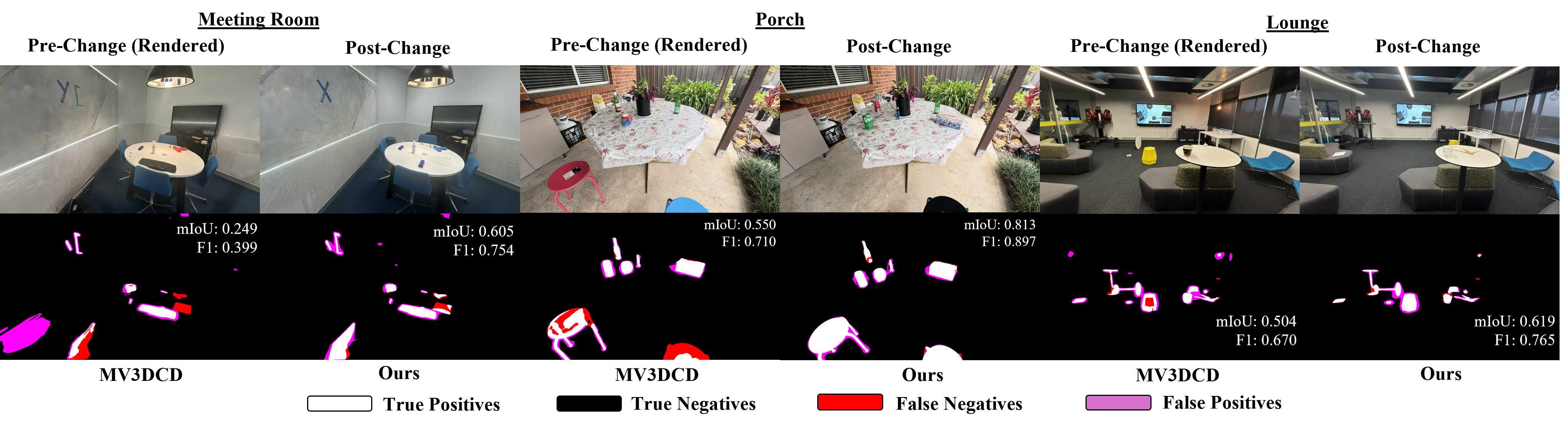}
   \vspace{-20pt}
   \caption{Qualitative comparison with MV3DCD~\cite{galappaththige2025multi}.
 MV3DCD's hard thresholding and intersection heuristic lead to missed or spurious detections, especially for subtle appearance changes in semantically similar objects (red-to-blue T-shaped object in Meeting Room, blue-to-black bench in Porch). Hard thresholding risks discarding subtle but important changes, while the intersection fails to capture true changes unless present in both masks. Our method jointly learns complementary change information in pixel- and feature-level cues via our novel self-supervised loss, capturing fine-grained changes and achieving state-of-the-art performance in both online and offline settings. 
}

   \vspace{-5pt}
   
   \label{fig:mv3dcd}
\end{figure*}
We introduce a novel SCD approach that, for the first time, unifies the strengths of \emph{online}, \emph{pose-agnostic}, and \emph{label-free} methods, while additionally enforcing \emph{multi-view consistency}~\cite{galappaththige2025multi} during inference. Our method surpasses existing online and offline methods in detection performance while operating at over 10~FPS. This leap in both speed and accuracy is enabled by two key algorithmic and system-level innovations: a novel self-supervised loss enforcing change information consistency across viewpoints, addressing the limitations of hard-thresholded intersection fusion~\cite{galappaththige2025multi} (see Fig.~\ref{fig:mv3dcd}),  and an ultra-light PnP-based pose estimation module. 

Our third innovation is a change-guided update strategy for the 3DGS-based scene representation.
Maintaining an up-to-date representation of an evolving scene without naively reconstructing it from scratch is challenging~\cite{ackermann2025clsplats,cai2023clnerf,zeng2025gaussianupdate,cheng2025lt}, but essential for long-term monitoring. Na\"ive reconstruction after each inspection round is computationally expensive and discards well-reconstructed information from unchanged regions. 
We address this by leveraging the predicted change masks to guide selective updates: only changed regions are newly reconstructed, fused with existing primitives, and refined through a lightweight global adjustment while preserving the geometry and the appearance of unchanged areas. Our selective update approach enables scene representation updates in seconds while reusing the high-fidelity representations of unchanged regions.

In summary, we make three contributions validated by our extensive experiments across real-world environments: 
\begin{itemize}
    \item We present an online approach for pose-agnostic SCD from unposed monocular images, operating in real time. Our approach is label-free and multi-view.
    \item We propose a novel self-supervised loss that jointly integrates feature- and pixel-level cues without heuristic fusion or hard-thresholding, achieving state-of-the-art performance in both online and offline settings.
    \item We introduce a change-guided selective reconstruction and fusion strategy that enables efficient, repeatable scene representation updates within seconds. 
\end{itemize}
\section{Related Work}
\label{sec:related_work}

\subsection{Scene Change Detection}
SCD has traditionally been studied as a bitemporal pairwise problem, where a model detects changes between two images captured at two time instances from identical~\cite{alcantarilla2018street,daudt2018fully,sakurada_change_2015,chen2021dr,varghese2018changenet,lei2020hierarchical} or closely-aligned~\cite{cyws2d,cyws3d,lin2025robust} viewpoints. Most approaches formulate this as a segmentation task~\cite{alcantarilla2018street,daudt2018fully,sakurada_change_2015,lin2025robust,chen2021dr}, while few explore bounding box prediction~\cite{cyws2d,cyws3d}, relying fully~\cite{alcantarilla2018street,daudt2018fully,sakurada_change_2015,chen2021dr,varghese2018changenet,lei2020hierarchical,cyws2d,cyws3d,lin2025robust} or partially~\cite{lee2024semi,sakurada2020weakly} on costly human annotations to compensate for lighting variations, seasonal changes, or viewpoint inconsistencies. However, this paradigm has clear limitations: performance degrades under distribution shifts, annotation is tedious, and the range of possible changes in complex scenes is virtually unbounded. Recent work~\cite{galappaththige2025multi,kim2025towards,cho2025zero,kannanZero,alpherts2025emplace} has shifted toward label-free or zero-shot approaches, driven by the emergence of powerful visual foundation models~\cite{kirillov2023segment,oquab2023dinov2}. 
However, these methods~\cite{kim2025towards, cho2025zero, alpherts2025emplace, kannanZero} assume image pairs with identical viewpoints are available---a condition rarely satisfied in autonomous systems operating along independent trajectories, and a condition we do not assume.

More recent methods~\cite{galappaththige2025multi,lu20253dgs,jiang2025gaussian} exploit high-fidelity 3D scene representations~\cite{kerbl20233d,wu20244d} to model the pre-change scene and render novel views from post-change viewpoints, motivating pose-agnostic SCD. MV3DCD~\cite{galappaththige2025multi} showed that learning change information across multiple viewpoints with a scene representation significantly outperforms pairwise predictions. However, these approaches require a complete set of pre- and post-change captures, and use Structure-from-Motion~\cite{schonberger2016structure} (SfM) to register poses to a common reference frame, confining them to an offline setting. In contrast, we study SCD in an online and incremental regime, where changes are inferred on-the-fly.

Pose-agnostic anomaly detection~\cite{zhou2023pad,kruse2024splatpose,liu2024splatpose+} also builds a 3D representation of a pre-change object. To detect anomalies, the object is rendered from the post-change viewpoint, then scored using feature comparisons. However, these works focus on single objects rather than large-scale scenes. Approaches such as~\cite{kruse2024splatpose,zhou2023pad} optimize camera poses directly against the representation, leading to slower pose estimation and, as shown by MV3DCD~\cite{galappaththige2025multi}, frequent convergence failures in large complex scenes with multiple changes and view-dependent inconsistencies. Liu \etal~\cite{liu2024splatpose+} improved efficiency by replacing this step with HLoc~\cite{sarlin2019coarse}.

While also employing a self-supervised objective, the approach of Furukawa \etal~\cite{Furukawa2020SelfsupervisedSA} differs fundamentally from our formulation of the self-supervised loss. Their loss is designed to exclude high-error regions to facilitate 2D alignment, whereas ours instead integrates complementary yet potentially noisy change cues across modalities into a persistent 3D representation.

ChangeSim~\cite{park2021changesim} also investigates online SCD. However, ChangeSim depends on an off-the-shelf RGB-D SLAM system~\cite{labbe2019rtab} for pose estimation and assumes that pre- and post-change trajectories are closely aligned. This reduces the task to image retrieval, where the nearest pre-change view (by \(L_1\) distance between camera poses) is selected. In contrast, we make no assumptions about incoming RGB-only frames or the trajectories; instead, we estimate poses directly in the pre-change coordinate frame and infer change masks by jointly leveraging all viewpoints observed so far. Our approach operates fully \textit{label-free}, \textit{pose-agnostic}, \textit{multi-view}, \textit{online}, and at \textit{real-time} rates.

\subsection{Efficient Representation Update}

NeRFs~\cite{mildenhall2021nerf} and 3DGS~\cite{kerbl20233d} are widely adopted photorealistic scene representations, capturing fine geometry and appearance. NeRFs regress a 5D plenoptic function~\cite{bergen1991plenoptic} using an MLP network to parameterize density and view-dependent radiance, while 3DGS employs anisotropic Gaussian primitives for real-time novel-view-synthesis. 

Recently, there has been growing interest in real-time reconstruction~\cite{meuleman2025fly,lin2025longsplat}. However, these methods generally underperform compared to offline counterparts and require substantial view overlap between frames. In contrast, approaches that address the long-term evolution of scenes remain less explored, focusing on updating representations from sparse and intermittent captures~\cite{ackermann2025clsplats,wu2023cl}. Closely related is continual learning for photorealistic scene representations~\cite{ackermann2025clsplats,zeng2025gaussianupdate,cai2023clnerf,wu2023cl}. NeRF-based continual learning approaches~\cite{wu2023cl,cai2023clnerf} utilize distillation or generative replay but inherit slow inference and longer optimization times. GaussianUpdate~\cite{zeng2025gaussianupdate} proposes a three-stage optimization pipeline requiring substantial training iterations. CL-Splats~\cite{ackermann2025clsplats} introduces a local optimization kernel to calculate gradients only for changed primitives, yet cannot robustly handle global appearance variations, particularly the illumination shifts often present between real-world inspection scenarios.

Long-term scene representations are also well-explored in robotics for autonomous navigation. To maintain spatial consistency in dynamic environments, these systems utilize volumetric representations~\cite{schmid2022panoptic,fu2022planesdf}, object-aware tracking~\cite{qian2023pov,zhu2024living}, unified metric-semantic frameworks~\cite{schmid2024khronos}, and recently, 3DGS~\cite{yugay2025gaussian}. 

Continual learning approaches~\cite{ackermann2025clsplats,zeng2025gaussianupdate,wu2023cl,cai2023clnerf} focus on updating representations while facilitating history recovery. In contrast, we focus on updating the representation with minimal training overhead, enabling frequent repeated inspections. To this end, we propose a simple selective modeling strategy that only reconstructs changed regions, guided by our change masks and the change representation, followed by fusion with existing primitives. A lightweight global optimization step ensures consistency, enabling updates within seconds while robustly handling both geometric and appearance changes, including global illumination variations.

\section{Methodology}

\begin{figure*}[t]
  \centering
   \includegraphics[width=\linewidth]{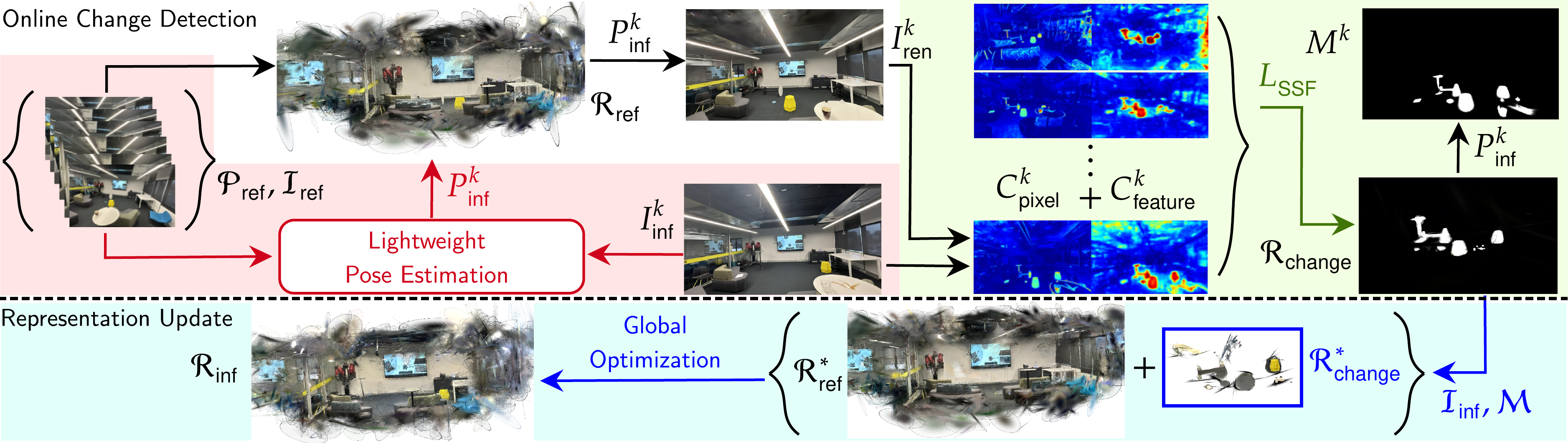}

   \caption{Proposed approach with this paper's contributions highlighted. We register an incoming inference image $I_{\text{inf}}^k$ to an existing reference representation $\mathcal{R}_{\text{ref}}$ with a \textcolor{red}{lightweight PnP-based pose estimator}. Using the estimated pose $P_{\text{inf}}^k$ and $\mathcal{R}_{\text{ref}}$ to render an aligned image $I_{\text{ren}}^k$, we extract change cues $C^k$ as a combination of pixel- and feature-level cues. \textcolor{ForestGreen}{Our novel self-supervised fusion loss $L_{\text{SSF}}$} guides the fusion of all observed change cues to build a change representation $\mathcal{R}_{\text{change}}$ that collectively learns change information from multiple viewpoints and infer change masks  $M^k$. Finally, we \textcolor{blue}{selectively reconstruct changed regions} to update the representation to $\mathcal{R}_{\text{inf}}$.}

   \vspace{-5pt}
   
   \label{fig:method}
\end{figure*}

Our approach is illustrated in Fig.~\ref{fig:method}. We begin by constructing a 3DGS~\cite{kerbl20233d} representation of the pre-change (\textit{reference}) scene offline (Sec.~\ref{sec:ref_rep}). Incoming images of the post-change (\textit{inference} scene) are processed online. We estimate its pose relative to the reference scene (Sec.~\ref{sec:pose}), then render the corresponding viewpoint to extract change cues (Sec.~\ref{sec:cues}). These cues are used to infer a change mask (Sec.~\ref{sec:infer}), leveraging current and previous observations. After processing all observations, the representation is updated (Sec.~\ref{sec:update}) to the current state of the environment.

\noindent \textbf{Problem Setup:} 
The reference scene \(\mathcal{R}_{\text{ref}}\) is captured with a set of \(n_{\text{ref}}\) images, \(\mathcal{I}_{\text{ref}} = \{I_{\text{ref}}^k\}_{k=1}^{n_{\text{ref}}}\). Over time, the scene undergoes changes in structure (e.g., additions, removals, or object movement), or object appearance (e.g., variations in color or texture), forming the inference scene \(\mathcal{R}_{\text{inf}}\). In addition, there may be `distractors' such as reflections, shadows and global illumination changes.

Our objective is to generate a binary change segmentation mask \(M^{k}\) for each incoming inference frame \(I_{\text{inf}}^{k}\) that localizes all relevant changes between \(\mathcal{R}_{\text{ref}}\) and \(\mathcal{R}_{\text{inf}}\) while suppressing distractors. After processing all inference frames \(\mathcal{I}_{\text{inf}} = \{I_{\text{inf}}^k\}_{k=1}^{n_{\text{inf}}}\), we obtain a set of refined masks \(\mathcal{M} = \{M^{k}_{\text{refined}}\}_{k=1}^{n_{\text{inf}}}\). Using the change masks \(\mathcal{M}\) together with \(\mathcal{I}_{\text{inf}}\), we selectively update the reference representation \(\mathcal{R}_{\text{ref}}\) to reflect the scene's current state \(\mathcal{R}_{\text{inf}}\).

\subsection{Building Reference Scene Representation}
\label{sec:ref_rep}

Following the standard 3DGS pipeline~\cite{kerbl20233d}, we first estimate the camera poses \(\mathcal{P}_{\text{ref}} = \{P_{\text{ref}}^k\}_{k=1}^{n_{\text{ref}}}\) for all images in \(\mathcal{I}_{\text{ref}}\) using SfM~\cite{schonberger2016structure}. Using \(\mathcal{I}_{\text{ref}}\), \(\mathcal{P}_{\text{ref}}\) and a sparse point cloud from SfM, we construct the \(\mathcal{R}_{\text{ref}}\) with Speedy-Splat~\cite{HansonSpeedy}. We assume that the viewpoints, scene coverage, and image quality in \(\mathcal{I}_{\text{ref}}\) are sufficient to produce a high-fidelity \(\mathcal{R}_{\text{ref}}\).

\subsection{Inferring Pose of an Incoming Frame}
\label{sec:pose}

For each reference image \(I_{\text{ref}}^{k} \in \mathcal{I}_{\text{ref}}\), we extract keypoints and descriptors using XFeat~\cite{potje2024xfeat} as a fast, lightweight detector, followed by exhaustive matching across all reference images. Using the known camera poses \(\mathcal{P}_{\text{ref}}\), these correspondences are triangulated to form a consistent 3D point set associated with each \(I_{\text{ref}}^{k}\). This point set serves as a geometric anchor for estimating the pose of incoming inference frames by establishing 2D--3D correspondences between their detected keypoints and the reference points. 

Given an incoming frame \(I_{\text{inf}}^{k}\), we extract its descriptors and select the top-\(n\) reference frames with the highest number of matches (\(n=4\) in our experiments). These reference frames provide candidate 2D--3D correspondences, which are then used to estimate the pose \(P_{\text{inf}}^{k}\) of \(I_{\text{inf}}^{k}\) via PnP with RANSAC~\cite{lepetit2009epnp,fischler1981ransac}. Finally, we refine \(P_{\text{inf}}^{k}\) with inliers through a GPU-parallel miniBA~\cite{meuleman2025fly}.

Since pose estimation for $I_{\text{inf}}^{k}$ relies exclusively on the retrieved reference frames, the system operates without drift accumulation. Moreover, by restricting inference to a fixed-size set of reference frames, we achieve constant-time $O(1)$ pose estimation. We discuss some limitations of the XFeat features in the supplementary material.

\subsection{Extracting Change Cues}
\label{sec:cues}

With \(P_{\text{inf}}^{k}\) expressed in the coordinate frame of \(\mathcal{P}_{\text{ref}}\), we query \(\mathcal{R}_{\text{ref}}\) to render the corresponding pre-change view \(I_{\text{ren}}^{k}\), matching the viewpoint of the incoming frame \(I_{\text{inf}}^{k}\). For the image pair \((I_{\text{inf}}^{k}, I_{\text{ren}}^{k})\), we extract change cues by computing differences at both the pixel and feature levels, capturing both appearance and structural changes.

\noindent\textbf{Pixel-level change cues:}  We quantify the differences between \((I_{\text{inf}}^{k}, I_{\text{ren}}^{k})\) at the pixel level using a combination of \(L_1\) and D-SSIM terms, following the photometric error formulation in 3DGS~\cite{kerbl20233d} (Eq.~\ref{eq:photo}) with \(\lambda=0.2\) following 3DGS~\cite{kerbl20233d}. We normalize \(C_{\text{pixel}}^k\) to \([0,1]\).
\begin{equation}
    \label{eq:photo}
    C_{\text{pixel}}^k = (1-\lambda)L_1+\lambda L_{\text{D-SSIM}}.
\end{equation}

\noindent\textbf{Feature-level change cues:} To capture high-level semantic differences, we leverage the visual foundation model SAM2-Tiny~\cite{ravi2024sam} to extract dense feature maps \((f_{\text{inf}}^{k}, f_{\text{ren}}^{k})\) for \((I_{\text{inf}}^{k}, I_{\text{ren}}^{k})\). Each feature map is represented as \(f \in \mathbb{R}^{\frac{h}{s} \times \frac{w}{s} \times d}\), where \(h\) and \(w\) denote the image height and width, \(s\) is the patch size, and \(d\) is the feature dimensionality. The feature-level change cues \(C_{\text{feature}}^{k}\) are computed as the absolute difference between the two feature maps:
\begin{equation}
    \label{eq:feature}
    C_{\text{feature}}^{k} = \sum_{i=1}^{d}  \left| f_{\text{inf}}^{k,i} - f_{\text{ren}}^{k,i} \right|\in \mathbb{R}^{\frac{h}{s} \times \frac{w}{s}},
\end{equation}
followed by bilinear interpolation to the original image resolution \((h, w)\) and normalization to the range \([0, 1]\).

\noindent\textbf{Combined change cues:}  
The final change cue map \(C^{k}\) for each \(I^k_{\text{inf}}\) combines pixel- and feature-level cues through simple addition (\(C^{k} = C_{\text{pixel}}^{k} + C_{\text{feature}}^{k}\)) balancing low-level appearance differences with high-level semantic variations.

This formulation leverages the complementary strengths of pixel- and feature-level cues while avoiding the loss of change information. Pixel-level cues effectively capture fine-grained appearance differences, such as color variations between semantically similar objects, but tend to be more sensitive to distractor changes caused by shadows, reflections, or illumination shifts. Feature-level cues are more robust to these distractors yet may struggle to detect subtle differences within semantically similar regions. MV3DCD~\cite{galappaththige2025multi} relies on hard thresholding, which can disregard subtle but relevant changes that fall below predefined thresholds. Moreover, since MV3DCD fuses its structure and feature-aware masks through intersection, it may further lose valid change information not simultaneously captured by both masks. When combined with our novel self-supervised fusion loss (Sec.~\ref{sec:infer}), the proposed formulation jointly integrates information across all observed viewpoints, effectively suppressing inconsistent distractors while maintaining sensitivity to meaningful changes.

\subsection{Inferring Change Masks}
\label{sec:infer}

MV3DCD~\cite{galappaththige2025multi} first enforced multi-view consistency for SCD using 3DGS~\cite{kerbl20233d}. We depart from its hard-thresholded heuristic fusion and introduce a novel self-supervised loss that jointly infers multi-view consistent change masks from all observed cues at test time.

After the reference scene \(\mathcal{R}_{\text{ref}}\) is constructed (Sec.~\ref{sec:ref_rep}), we initialize the \emph{change} representation \(\mathcal{R}_{\text{change}}\) from \(\mathcal{R}_{\text{ref}}\) by discarding all color parameters and introducing a learnable change parameter \(c\)~\cite{galappaththige2025multi} for each primitive.

\(\mathcal{R}_{\text{change}}\) serves two purposes: (1) it enables fusing change cues \(C^k\) from any viewpoint into a single, multi-view consistent change representation, and (2) it acts as a persistent memory that carries change information over observing viewpoints. As a result, when a new frame arrives, the incoming change cues are fused with all previously observed cues in \(\mathcal{R}_{\text{change}}\). Rendering \(\mathcal{R}_{\text{change}}\) at the pose \(P^k_{\text{inf}}\) yields the predicted change mask \(M^k\) for that viewpoint.

For an incoming $I_{\text{inf}}^k$, before inferring the change mask \(M^k\), we update \(\mathcal{R}_{\text{change}}\) for $n$ iterations ($n=16$ in our experiments) using our self-supervised fusion loss:
\begin{equation}
\label{eq:loss}
L_{\text{SSF}} = C^{i} \odot (1 - \tilde{M}^{i}) 
    + \log\!\big(1 + \operatorname{mean}(\tilde{M}^{i})^{2}\big),
\end{equation}
\noindent where \(\odot\) denotes Hadamard (i.e. element-wise) multiplication and \(\tilde{M}^{i}\) is the sigmoid-activated rendered change mask \(\sigma(M_{\text{ren}}^{i})\) from the viewpoint of the $i$-th frame. At each iteration, we randomly sample $i$ from all past inference frame IDs $i \in [0, k]$, but biased towards the most recent frame $k$ with $1/3$ probability. $C^i$ contains the combined change cues of the $i$-th frame (Sec.~\ref{sec:cues}). We infer the change mask \(M^k\) for the $k^{\text{th}}$ frame after this optimization.

Intuitively, minimizing \(L_{\text{SSF}}\) encourages the change parameters $\tilde{c}$ in $\mathcal{R}_\text{change}$ to change so that the rendered \(\tilde{M}^{i}\) has values close to 1 in regions where change cues are strong via the term \(C^{i} \odot (1 - \tilde{M}^{i})\). To prevent the trivial solution of \(\tilde{M}^{i} = 1\) everywhere, the regularization term \(\log\!\big(1 + \operatorname{mean}(\tilde{M}^{i})^{2}\big)\) is included. 

This formulation allows us to infer \(M^k\) for $I_{\text{inf}}^k$ jointly from all past and current change cues. By accumulating change information from all observed frames in $\mathcal{R}_\text{change}$, we enforce multi-view consistency and mitigate view-dependent distractors from irrelevant changes.

\subsection{Scene Representation Update}
\label{sec:update}

After completing online change detection for all observations, we perform a post-refinement of \(\mathcal{R}_{\text{change}}\) using 
all \(C^k\), and render refined change masks \(M^k_{\text{refined}}\) for \(k \in [0, n_{\text{inf}}]\).  

We then discard \(c\) from each primitive in \(\mathcal{R}_{\text{change}}\) and introduce view-dependent appearance modeled via spherical harmonics~\cite{kerbl20233d}. To only reconstruct changed regions, we mask the inference images using the refined change masks as \(\hat{I}^k_{\text{inf}} = I^k_{\text{inf}} \odot M^k_{\text{refined}}\). \(\hat{I}^k_{\text{inf}}\) guides the reconstruction of changed regions \(\mathcal{R}_{\text{change}}^*\) following the standard 3DGS~\cite{kerbl20233d} optimization pipeline. This disentangled reconstruction is highly efficient and requires a fraction of the primitives compared to modeling the entire scene, thereby accelerating rendering and avoiding redundant computations in unchanged regions. Notably, this selective reconstruction achieves rendering speeds exceeding 400 FPS, substantially reducing overall optimization time.

Next, we fuse \([\mathcal{R}^*_{\text{ref}}, \mathcal{R}_{\text{change}}^*]\) to form the inference scene \(\mathcal{R}_{\text{inf}}\), where \(\mathcal{R}^*_{\text{ref}}\) denotes \(\mathcal{R}_{\text{ref}}\) excluding primitives that contribute to changed pixels. A fast 
global optimization is then performed, guided by \(\mathcal{I}_{\text{inf}}\). We restrict the adaptive density control~\cite{kerbl20233d} only to the primitives contributing to changed pixels in at least one view to avoid unnecessary densification in unchanged regions.

This restricted global refinement serves multiple purposes: (1) it accounts for global illumination differences, (2) it mitigates boundary artifacts that may arise around the changed regions after fusion, and (3) it corrects residual errors due to imperfect change masks. Our design reuses primitives from \(\mathcal{R}_{\text{ref}}\) wherever possible, while \(\mathcal{R}^*_{\text{change}}\) efficiently models new structures. Together, these design choices significantly speed up optimization, enabling complete scene updates within seconds.

\definecolor{offlineColor}{RGB}{255, 253, 150} 
\definecolor{onlineColor}{RGB}{233, 255, 243}  
\definecolor{bestColor}{RGB}{255,200,200}       
\definecolor{secondbestColor}{RGB}{255, 235, 235}  
\newcommand{\cmark}{\ding{51}} 

\newcommand{\bestcell}[1]{\textbf{#1}}
\newcommand{\secondbestcell}[1]{\textit{#1}}


\section{Experiments}
\label{sec:exp}

\noindent\textbf{Datasets:} 
We evaluate our method on PASLCD~\cite{galappaththige2025multi} for SCD. PASLCD comprises 10 room-scale (i.e., a cantina) indoor and outdoor scenes captured under similar and varying lighting conditions. It features both surface-level appearance and object-level geometric changes, along with numerous distractors such as shadows, reflections, and illumination shifts, making it a highly challenging multi-view dataset for SCD. Importantly, PASLCD captures scenes from unconstrained, independently traversed camera trajectories, closely reflecting real-world autonomous operation. 
For the scene representation update, we evaluate on PASLCD~\cite{galappaththige2025multi} and CL-Splats~\cite{ackermann2025clsplats}. CL-Splats consists of five small-scale (i.e., tabletop) scenes, each featuring a single object-level change.

\begin{table}[t]
\small
\centering
\caption{
Quantitative results for SCD on PASLCD~\cite{galappaththige2025multi} averaged over all 20 instances.
LF: Label-Free, PA: Pose-Agnostic, MV: Multi-View consistency for change detection, ON: Online.
We additionally report the total runtime, including pose estimation and reference reconstruction, for \colorbox{offlineColor}{offline} methods, and the operating frame rate (FPS) for \colorbox{onlineColor}{online} methods.
Our method achieves the best performance in both settings, even outperforming all existing offline methods while operating online.
}

\setlength{\tabcolsep}{1.5pt}
\renewcommand{\arraystretch}{1.}
\begin{tabular}{lccccccc}
\toprule
\textbf{Method} & \textbf{LF} & \textbf{PA} & \textbf{MV} & \textbf{ON} & \textbf{mIoU} & \textbf{F1} & {\footnotesize \makecell{\colorbox{offlineColor}{\textbf{Runtime}}\\/\colorbox{onlineColor}{\textbf{FPS}}}}  \\ 
\midrule
\rowcolor{offlineColor} R-SCD~\cite{lin2025robust} & -- & -- & -- & -- & 0.118 & 0.199 & 194s \\
\rowcolor{offlineColor} CYWS2D~\cite{cyws2d} & -- & -- & -- & -- & 0.273 & 0.398 & \secondbestcell{189s}\\
\rowcolor{offlineColor} GeSCD~\cite{kim2025towards} & \cmark & -- & -- & -- & 0.477 & 0.611 & 298s \\
\rowcolor{offlineColor} ZeroSCD~\cite{cho2025zero} & \cmark & -- & -- & -- & 0.306 & 0.414 & 409s \\
\rowcolor{offlineColor} 3DGS-CD~\cite{lu20253dgs} & \cmark & \cmark & \cmark & -- & 0.209 & 0.339 & 824s\\
\rowcolor{offlineColor} MV3DCD~\cite{galappaththige2025multi} & \cmark & \cmark & \cmark & -- & \secondbestcell{0.478} & \secondbestcell{0.628} & 479s \\
\rowcolor{offlineColor} \textbf{Ours \footnotesize{(Offline)}} & \cmark & \cmark & \cmark & -- & \bestcell{0.552} & \bestcell{0.694} & \bestcell{156s} \\
\midrule
\rowcolor{onlineColor} ChangeSim (CS)~\cite{park2021changesim} & -- & -- & -- & \cmark & 0.018 & 0.034 &   \bestcell{11.5} \\
\rowcolor{onlineColor} CS+CYWS2D~\cite{cyws2d} & -- & -- & -- & \cmark & \secondbestcell{0.243} & \secondbestcell{0.360} & 8.2 \\
\rowcolor{onlineColor} CS+GeSCD~\cite{kim2025towards} & \cmark & -- & -- & \cmark & 0.181 & 0.270 & \textless 1 \\
\rowcolor{onlineColor} OmniposeAD~\cite{zhou2023pad} & \cmark & \cmark & -- & \cmark & 0.168 & 0.262 & \textless 1 \\
\rowcolor{onlineColor} SplatPose~\cite{kruse2024splatpose} & \cmark & \cmark & -- & \cmark & 0.173 & 0.281 & \textless 1 \\
\rowcolor{onlineColor} SplatPose+~\cite{liu2024splatpose+} & \cmark & \cmark & -- & \cmark & 0.237 & 0.358 & \textless 1 \\
\rowcolor{onlineColor} \textbf{Ours} & \cmark & \cmark & \cmark & \cmark & \bestcell{0.486} & \bestcell{0.638} & \secondbestcell{11.2} \\
\bottomrule
\end{tabular}
\label{tab:scd}
\end{table}

\noindent\textbf{Baselines and Metrics:} 
We conduct a comprehensive evaluation using the best-performing baselines~\cite{lin2025robust,cyws2d,kim2025towards,cho2025zero,lu20253dgs,galappaththige2025multi,park2021changesim,zhou2023pad,kruse2024splatpose,liu2024splatpose+}. For pairwise methods~\cite{lin2025robust,cyws2d,kim2025towards,cho2025zero,lu20253dgs} evaluated in the offline setting, we render identical viewpoints using vanilla 3DGS~\cite{kerbl20233d} for a fair comparison, although this substantially simplifies the task by removing viewpoint inconsistencies. 
For the online setting, we construct two additional baselines by integrating ChangeSim's~\cite{park2021changesim} frame matching with the best-performing pairwise methods~\cite{cyws2d,kim2025towards}. We provide ground-truth poses for ChangeSim's frame retrieval module to ensure a fair evaluation, as PASLCD lacks depths for ChangeSim's off-the-shelf RGB-D SLAM system. We use model checkpoints provided by the authors for the supervised methods~\cite{lin2025robust,cyws2d}.
Following standard practice in SCD~\cite{galappaththige2025multi,lin2025robust,alcantarilla2018street,sakurada2020weakly}, we report mean intersection over union (mIoU) and F1 score computed for change pixels in the ground-truth mask.

For efficient scene representation update, we adopt 3DGS and fast variants~\cite{HansonSpeedy,hoellein_2025_3dgslm,kerbl20233d}  as our baselines. We also evaluate CLNeRF~\cite{cai2023clnerf} among publicly available continual learning methods. 
Following standard evaluation protocols~\cite{cai2023clnerf,ackermann2025clsplats,zeng2025gaussianupdate,wu2023cl,kerbl20233d}, we report PSNR, SSIM, and LPIPS for novel views after scene update, along with runtimes.

\subsection{Experiments on Scene Change Detection}

\noindent \textbf{Offline SCD Results:} 
Table~\ref{tab:scd} presents an extensive comparison against SOTA methods across all SCD settings on PASLCD~\cite{galappaththige2025multi}. 
In the offline setting, we follow the protocol of MV3DCD~\cite{galappaththige2025multi} by optimizing \(\mathcal{R}_{\text{change}}\) using our $L_{\text{SSF}}$ with access to all inference views jointly for 3k iterations. 

Generally, label-free approaches yield better performance. 
Among these, our method achieves the highest overall performance, improving mIoU by approximately \textbf{15\%} over the strongest offline competitor, MV3DCD, while running nearly \textbf{3$\times$ faster}. 
The SCD performance gain primarily stems from our proposed self-supervised fusion loss \(L_{\text{SSF}}\), which eliminates the hard thresholding and intersection heuristics used in MV3DCD, enabling more robust and fine-grained change localization.
\begin{table}[t]
\small
\centering
\caption{
Runtime analysis of each module in our online SCD pipeline, measured in milliseconds per frame on PASLCD~\cite{galappaththige2025multi}. 
Most of the computation time is spent on multi-view change information fusion, while other modules are lightweight.
}
\setlength{\tabcolsep}{4pt}
\renewcommand{\arraystretch}{1.}
\begin{tabular}{lcc}
\toprule
\textbf{Module} & \textbf{ms/Frame} & \textbf{Percentage (\%)} \\
\midrule
Extracting Descriptors & 1.28 & 1.4 \\
Reference Image Retrieval & 11.50 & 12.8 \\ 
Pose Estimation & 16.47 & 18.4 \\ 
Change Cue Generation & 1.69 & 1.9 \\
Multi-View Change Cue Fusion & 58.17 & 64.9 \\ 
Change Mask Inference & 0.49 & 0.6 \\
\midrule
Total & 89.60 & 100 \\
\bottomrule
\end{tabular}
\label{tab:runtime}
\end{table}

\begin{figure}[b]
  \centering
  \scalebox{0.75}{
  \begin{tikzpicture}
    \begin{axis}[
      width=0.9\linewidth,
      xlabel={Number of Optimization Iterations},
      ylabel={FPS},
      ymin=4, ymax=21,
      ytick={4,8,12,16,20},
      xtick={2,4,8,16,32},      
      grid=major,
       xmajorgrids=true,  
      ymajorgrids=false,   
      axis y line*=left,
      axis x line*=bottom,
      legend style={
        at={(0.97,0.03)},       
        anchor=south east,
        legend columns=2,
        column sep=8pt,
        draw=black, fill=white, rounded corners,
        font=\footnotesize
      },
    ]
      \addplot[mark=square*, thick, blue] coordinates {
        (2,19.8) (3,18.3) (4,17.2) (8,13.3) (16,11.1) (32,7.8)
      };
      \addlegendentry{FPS}

      \addlegendimage{line legend, red, dashed, mark=*}
      \addlegendentry{F1 Score}
    \end{axis}

    \begin{axis}[
      width=0.9\linewidth,
      ylabel={F1 Score},
      ymin=0.61, ymax=0.642,
      ytick={0.61,0.62,0.63,0.64},
      xtick={2,4,8,16,32},
      axis y line*=right,
      axis x line=none,
    ]
      \addplot[mark=*, thick, red, dashed, forget plot] coordinates {
        (2,0.6149) (3,0.6222) (4,0.6310) (8,0.6333) (16,0.6380) (32,0.6381)
      };
    \end{axis}
  \end{tikzpicture}
}
\vspace{-5pt}
  \caption{Speed–accuracy trade-off of our online method. Our method can operate between 11--20 FPS with a relative performance drop of 3.6\% in F1 Score.}
  \label{fig:speed_miou}
\end{figure}
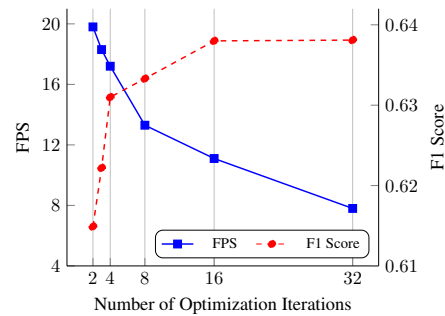

\begin{figure*}[t]
  \centering
   \includegraphics[width=\linewidth]{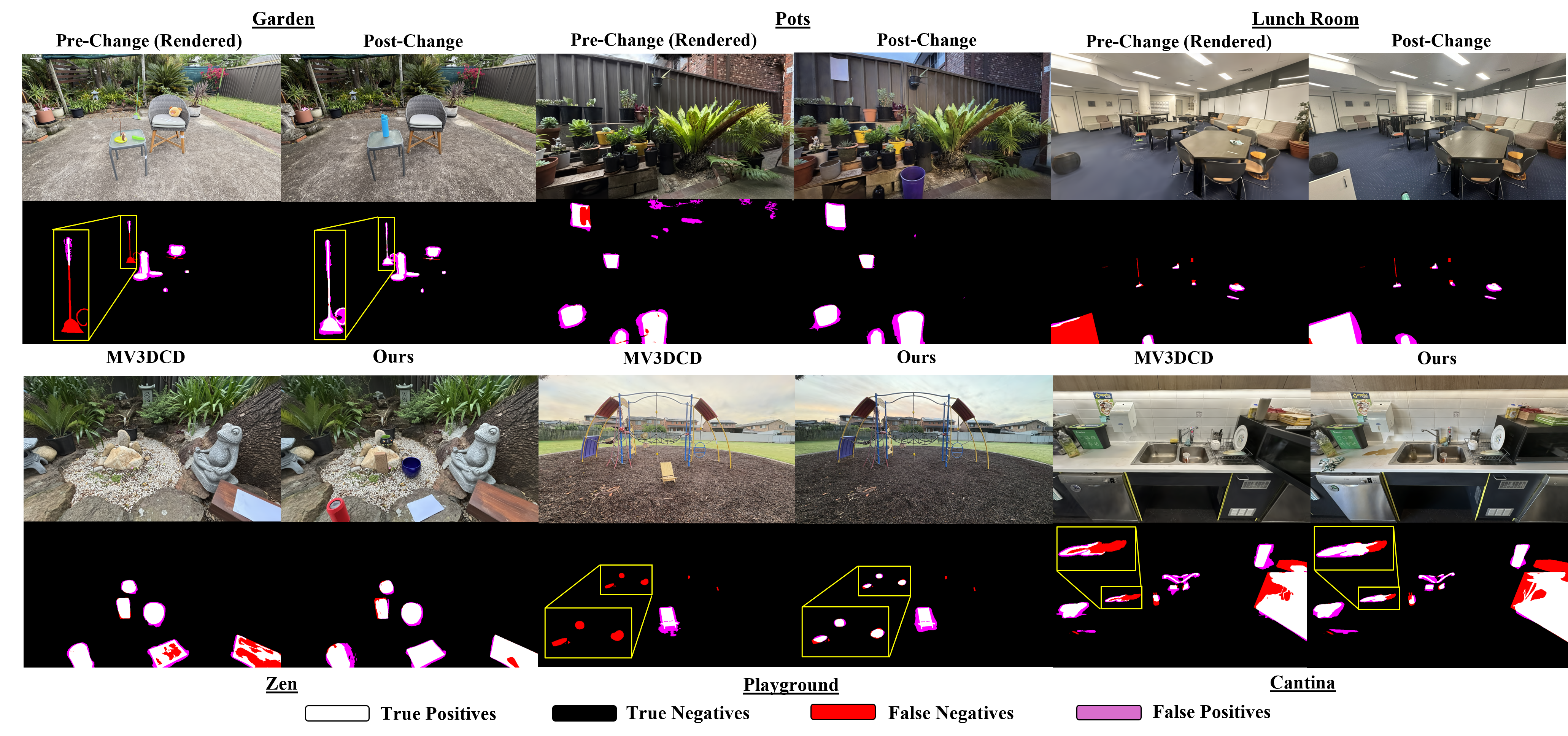}

  \caption{Additional qualitative comparison with MV3DCD~\cite{galappaththige2025multi}.
Our rendered change masks align more closely with the ground truth, capturing subtle structural and appearance changes that MV3DCD often misses. 
In contrast to MV3DCD, our method produces fewer spurious detections and demonstrates strong robustness to distractor variations across both indoor and outdoor environments.}
   \vspace{-10pt}
   
   \label{fig:scd_vis}
\end{figure*}

\noindent \textbf{Online SCD Results:}
We compare our approach with existing online SCD methods (Table~\ref{tab:scd}). 
In the online setting, our approach achieves \textbf{2$\times$ higher mIoU} than the strongest competitor, CS+CYWS2D~\cite{cyws2d}, while maintaining real-time performance at \textbf{11~FPS}. 
Notably, SOTA offline methods such as GeSCD~\cite{kim2025towards} experience severe degradation when exposed to viewpoint discrepancies. 
CYWS2D exhibits a smaller performance drop under such conditions, likely due to its pretraining on COCO-Inpainted~\cite{cyws2d}, which includes image pairs with viewpoint variations. 
The upper bounds of these two methods under identical viewpoints are shown in the offline setting. 
Remarkably, our online approach not only establishes SOTA results among online methods but also surpasses the best offline models, demonstrating strong robustness and efficiency under real-world conditions. 
We conduct all experiments at $1008\times560$ resolution on a GeForce RTX4090. FPS was measured as empirical wall-clock time over the number of frames processed.

\noindent \textbf{Runtime Analysis:} Table~\ref{tab:runtime} summarizes the runtime breakdown (asynchronous overhead) of our online SCD pipeline for a single inference frame \(I^k_{\text{inf}}\). 
Approximately 33\% of the total time is spent on pose estimation, aligning \(I^k_{\text{inf}}\) to the coordinate system of \(\mathcal{R}_{\text{ref}}\). 
SCD accounts for 67\% of the overall runtime, of which the majority (94\%) is attributed to multi-view change cue fusion spent on optimizing \(\mathcal{R}_{\text{change}}\).

\noindent \textbf{Speed Vs. Accuracy Trade-off: } We vary the number of iterations used for multi-view change cue fusion---the dominant contributor to runtime (Fig.~\ref{fig:speed_miou}). Our approach can operate between 11--20~FPS, with only a modest 3.6\% drop in F1 score. This efficiency stems from $\mathcal{R}_{\text{change}}$, which serves as a persistent memory of previously learned changes, minimizing the iterations needed for subsequent frames.

\noindent \textbf{Qualitative Results:} 
We present qualitative comparisons against our closest competitor, MV3DCD~\cite{galappaththige2025multi}, in Figs.~\ref{fig:mv3dcd} and~\ref{fig:scd_vis}. 
Our method demonstrates superior change localization and robustness across diverse real-world scenes. 
Unlike MV3DCD, our approach produces cleaner and more spatially coherent change masks with significantly fewer false negatives. 
It effectively captures subtle appearance and geometric changes that MV3DCD often overlooks, while suppressing false positives caused by distractors.
These results highlight that our self-supervised fusion of pixel- and feature-level cues combined with multi-view consistency enables accurate, fine-grained change detection.
\begin{table}[t]
\small
\centering
\caption{
Ablation study of our SCD approach on PASLCD~\cite{galappaththige2025multi}. Performance benefits from every component.
}
\setlength{\tabcolsep}{4pt}
\renewcommand{\arraystretch}{1.}
\begin{tabular}{@{}lcc@{}}
\toprule
\textbf{Variant} & \textbf{mIoU~$\uparrow$} & \textbf{F1~$\uparrow$} \\
\midrule
\textbf{Ours (Full)} & \textbf{0.486} & \textbf{0.638} \\
-- $L_1$ & 0.320 & 0.464 \\
-- $L_{\text{D-SSIM}}$ & 0.447 & 0.620 \\
-- $C_{\text{pixel}}$ (using only $C_{\text{feature}}$) & \ding{55} & \ding{55} \\
-- $C_{\text{feature}}$ (using only $C_{\text{pixel}}$) & \ding{55} & \ding{55} \\
-- Regularization term & \ding{55} & \ding{55} \\
\midrule
Ours \footnotesize{(With \cite{galappaththige2025multi}'s Thresholding \& Heuristic Fusion)} & 0.350 &  0.495 \\

\bottomrule
\end{tabular}
\label{tab:ablation_loss}
\end{table}

\begin{table*}[t]
\small
\centering
\caption{
Quantitative comparison of scene representation update on PASLCD~\cite{galappaththige2025multi} and CL-Splats~\cite{ackermann2025clsplats}. 
Our method achieves comparable or higher reconstruction quality than approaches that fully re-optimize the evolved scene from scratch, 
while providing updated representations within seconds ($<60$s), achieving up to \textbf{8--9$\times$ faster} runtimes. Results are averaged over all instances and scenes.
}

\setlength{\tabcolsep}{4pt}
\renewcommand{\arraystretch}{1.}
\begin{tabular}{lcccccccc}
\toprule
\multirow{2}{*}{\textbf{Method}} & 
\multicolumn{4}{c}{\textbf{PASLCD}} & 
\multicolumn{4}{c}{\textbf{CL-Splats}} \\
\cmidrule(lr){2-5} \cmidrule(lr){6-9}
 & PSNR (dB)~$\uparrow$ & SSIM~$\uparrow$ & LPIPS~$\downarrow$ & Runtime (s)~$\downarrow$ & 
 PSNR (dB)~$\uparrow$ & SSIM~$\uparrow$ & LPIPS~$\downarrow$ & Runtime (s)~$\downarrow$ \\
\midrule
3DGS~\cite{kerbl20233d} & 22.21 & 0.7558 & \secondbestcell{0.2426} & 550 & \secondbestcell{30.31 }& 0.9319 & \secondbestcell{0.1178}  & 364 \\
3DGS-LM~\cite{hoellein_2025_3dgslm} & \secondbestcell{22.26 }& 0.7562 & \bestcell{0.2422} & \secondbestcell{340} & 29.95 & 0.9322 & \bestcell{0.1177}& \secondbestcell{275} \\
SpeedySplats~\cite{HansonSpeedy} & 22.25 & \secondbestcell{0.7603} & 0.2618 & 399 & 29.89 & \secondbestcell{0.9349} & 0.1290 & 312 \\
\midrule
CLNeRF~\cite{cai2023clnerf} & 22.27 & 0.6239 & 0.3907 & 451  & 26.29 & 0.7867 & 0.2235 &  301 \\
\textbf{Ours} & \bestcell{23.70}  & \bestcell{0.7868} & 0.2491 & \bestcell{42} & \bestcell{30.54} &\bestcell{0.9356} & 0.1256 & \bestcell{39} \\
\bottomrule
\end{tabular}
\label{tab:sceneupdate}
\end{table*}

\begin{figure*}[t]
  \centering
   \includegraphics[width=\linewidth]{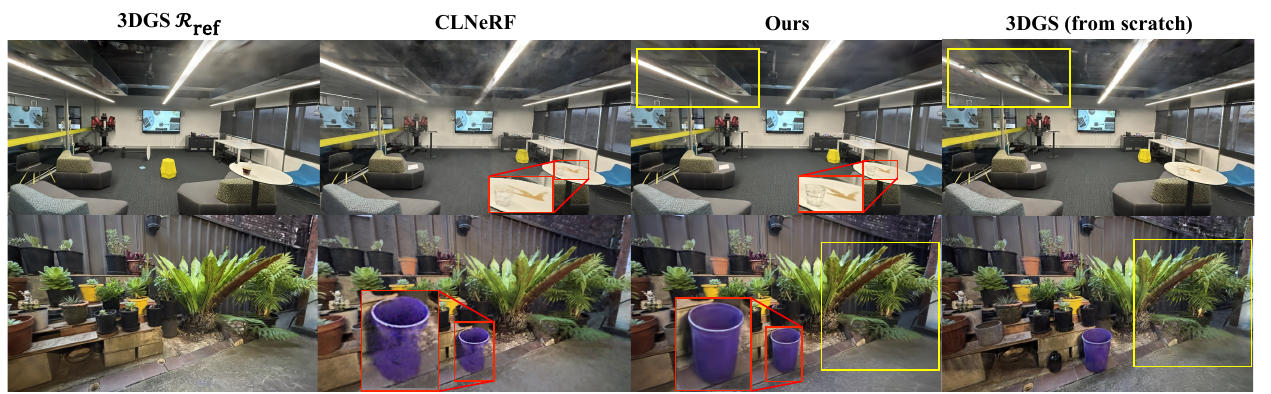}

\caption{
Qualitative comparison of rendered views from the updated representation with CLNeRF~\cite{cai2023clnerf} and 3DGS~\cite{kerbl20233d} (from scratch). 
Our method more accurately reconstructs changed regions (red boxes) while reusing primitives from \(\mathcal{R}_{\text{ref}}\) to preserve high fidelity in unchanged areas (yellow boxes), compared to na\"ive reconstruction at each time.
}

   \vspace{-5pt}
   
   \label{fig:update}
\end{figure*}

 \noindent \textbf{Ablation Analysis:} 
We conduct an ablation study on PASLCD~\cite{galappaththige2025multi}  (Table~\ref{tab:ablation_loss}). 
Removing either the $L_1$ or $L_{\text{D-SSIM}}$ term in $C_{\text{pixel}}$ noticeably degrades performance, with the former providing a relatively stronger supervisory signal for $L_{\text{SSF}}$. 
When using either $C_{\text{pixel}}$ or $C_{\text{feature}}$, the model fails to converge, indicating that neither modality alone sufficiently guides  $L_{\text{SSF}}$. 
Removing the regularization term collapses training into the trivial solution $\tilde{M}^{k} = 1$ (disscussed in sec.~\ref{sec:infer}).
Using MV3DCD's~\cite{galappaththige2025multi} hard thresholding and intersection heuristic instead of $L_\text{SSF}$ degrades performance.

\begin{table}[t]
\small
\centering
\caption{
Analysis on scene update component on PASLCD~\cite{galappaththige2025multi}. 
Runtime for change detection and refinement is excluded. GO: Global Optimization, SR: Selective Reconstruction.
}
\setlength{\tabcolsep}{1pt}
\renewcommand{\arraystretch}{1.0}
\begin{tabular}{lcccc}
\toprule
\textbf{Variant} & \textbf{PSNR (dB)~$\uparrow$} & \textbf{SSIM~$\uparrow$} & \textbf{LPIPS~$\downarrow$} &\textbf{Runtime (s)~$\downarrow$}\\

\midrule
GO Only (3DGS) & 22.64 & 0.7611 & 0.2550 & 145 \\
GO Only (Ours) & 23.01 & 0.7751 & 0.2553 & 79 \\

SR Only & 19.89 & 0.6814 & 0.3084 & \textbf{28} \\
\midrule
\textbf{Ours (Full)} & \textbf{23.70} & \textbf{0.7868} & \textbf{0.2491} &  36 \\
\bottomrule
\end{tabular}
\label{tab:ablation_update}
\end{table}
\subsection{Experiments on 3DGS Representation Update}

\noindent\textbf{Quantitative Results:} 
We compare our change-guided representation update strategy against reconstructing the scene from scratch~\cite{kerbl20233d,hoellein_2025_3dgslm, HansonSpeedy} and updating~\cite{cai2023clnerf} in Table~\ref{tab:sceneupdate}. 
Our method achieves comparable or slightly superior performance while substantially reducing the training overhead. 
For example, total optimization time is $\textbf{8}\times$ \textbf{faster} than 3DGS-LM~\cite{hoellein_2025_3dgslm} and \textbf{13}$\times$ \textbf{faster} than 3DGS~\cite{kerbl20233d} on PASLCD~\cite{galappaththige2025multi}. 
The slight performance gain arises from reusing the well-reconstructed, unchanged regions of the reference scene, which may not be well captured by the limited set of inference views.

\noindent\textbf{Qualitative Results:} 
We present qualitative comparisons with 3DGS~\cite{kerbl20233d} and CLNeRF~\cite{cai2023clnerf} in Fig.~\ref{fig:update}. 
Our method more accurately reconstructs the changed regions compared to CLNeRF, while also achieving higher visual fidelity in unchanged areas than 3DGS (built from scratch), owing to the effective reuse of primitives from the reference scene.

\noindent\textbf{Ablation Analysis:} 
To evaluate runtime efficiency (Table~\ref{tab:ablation_update}), all experiments are conducted for 10k iterations.
We begin with standard global optimization following 3DGS’s adaptive density control~\cite{kerbl20233d}. 
Restricting this process to primitives associated with changed pixels accelerates training by avoiding densification in unchanged regions. 
Selective reconstruction alone runs approximately \textbf{5$\times$ faster} than standard global optimization but introduces local artifacts when used in isolation (discussed in Sec.~\ref{sec:update}). 
Combining selective reconstruction with our global optimization achieves the best of both approaches, where the former efficiently models new geometry while the latter corrects residual artifacts locally and illumination differences globally.

\section{Conclusion}

We proposed a novel approach to pose-agnostic SCD that detects change in an online manner with SOTA performance, outperforming even the best offline methods. We introduce two key algorithmic and system-level innovations to achieve this: an ultra-light PnP-based pose estimator and a self-supervised fusion loss for learning a multi-view consistent change representation. Additionally, we introduced a change-guided update strategy for 3DGS, reducing training overhead to seconds while retaining reconstruction fidelity. Future work may focus on developing richer complementary change cues, potentially improving both online and offline SCD performance.

 \section{Acknowledgment }
This work was supported by the Australian Research Council Research Hub in Intelligent Robotic Systems for Real-Time Asset Management (ARIAM) (IH210100030) and Abyss Solutions. C.J., N.S., and D.M. also acknowledge ongoing support from the QUT Centre for Robotics.

{
    \small
    \bibliographystyle{ieeenat_fullname}
    \bibliography{main}
}


\end{document}